\documentclass{main}
\usepackage{lipsum} 

\usepackage{amsmath}
\usepackage[colorlinks=true, allcolors=blue]{hyperref}
\usepackage{multirow}
\usepackage[table,xcdraw]{xcolor}
\usepackage{wrapfig}
\usepackage{multicol}
\usepackage{ragged2e}
\usepackage{etoolbox}
\AtBeginEnvironment{multicols}{\RaggedRight}
\usepackage[superscript,biblabel,nomove]{cite}

\newcommand\blfootnote[1]{%
  \begingroup
  \renewcommand\thefootnote{}\footnote{#1}%
  \addtocounter{footnote}{-1}%
  \endgroup
}

\usepackage{todonotes}

\begin{document}

\title{Leveraging Large Language Models to Extract Information on Substance Use Disorder Severity from Clinical Notes: A Zero-shot Learning Approach}

\author{Maria Mahbub$^1$$^,$$^*$, Gregory M. Dams$^2$, Sudarshan Srinivasan$^1$, Caitlin Rizy$^1$, Ioana Danciu$^1$, Jodie Trafton$^2$, Kathryn Knight$^1$$^,$$^*$}

\institutes{
    $^1$Oak Ridge National Laboratory, Oak Ridge, TN;
    $^2$Program Evaluation and Resource Center, Office of Mental Health and Suicide Prevention, Department of Veterans Affairs, Menlo Park, CA
}

\maketitle

\section*{Abstract}
\vspace{-.5em}
\textit{Substance use disorder (SUD) poses a major concern due to its detrimental effects on health and society. SUD identification and treatment depend on a variety of factors such as severity, co-determinants (e.g., withdrawal symptoms), and social determinants of health. Existing diagnostic coding systems used by American insurance providers, like the International Classification of Diseases (ICD-10), lack granularity for certain diagnoses, but clinicians will add this granularity (as that found within the Diagnostic and Statistical Manual of Mental Disorders classification or DSM-5) as supplemental unstructured text in clinical notes. Traditional natural language processing (NLP) methods face limitations in accurately parsing such diverse clinical language. Large Language Models (LLMs) offer promise in overcoming these challenges by adapting to diverse language patterns. This study investigates the application of LLMs for extracting severity-related information for various SUD diagnoses from clinical notes. We propose a workflow employing zero-shot learning of LLMs with carefully crafted prompts and post-processing techniques. Through experimentation with Flan-T5, an open-source LLM, we demonstrate its superior recall compared to the rule-based approach. Focusing on 11 categories of SUD diagnoses, we show the effectiveness of LLMs in extracting severity information, contributing to improved risk assessment and treatment planning for SUD patients.
}

\section*{Introduction}
\vspace{-.5em}
Substance use disorder (SUD) is a multidimensional health problem growing at a national scale. The complexity of SUD assessment and treatment is based on several factors, including symptom expression, patterns of substance use over time, co-morbidities, and other social determinants of health impacting clinical presentation and treatment outcomes. Despite this complexity, the codes available for identifying patients with SUD, such as the International Classification of Diseases(ICD-10)\cite{ICD}, in clinical health records lack the granularity for indicating these additional factors, which makes it difficult to automatically flag any patients who may present specific risk factors. In addition to the ICD-10 coding system, the Diagnostic and Statistical Manual of Mental Disorders classification (DSM-5)\cite{DSM} is also used to set criteria for various disorders as defined by the medical community. However, the DSM-5 does not designate a numeric code for any of the conditions it lists (and therefore cannot easily enter into a structured electronic healthcare record like an ICD-10 code can), nor is there a one-to-one equivalency for a DSM-5 diagnosis and an ICD-10 code diagnosis.
\blfootnote{This manuscript has been authored by UT-Battelle, LLC, under contract DE-AC05-00OR22725 with the US Department of Energy (DOE). The US government retains and the publisher, by accepting the article for publication, acknowledges that the US government retains a nonexclusive, paid-up, irrevocable, worldwide license to publish or reproduce the published form of this manuscript, or allow others to do so, for US government purposes. DOE will provide public access to these results of federally sponsored research in accordance with the DOE Public Access Plan (\url{https://www.energy.gov/doe-public-access-plan}).}

When clinicians use DSM-5 to diagnose their patients and ICD-10 to codify this information, it is not uncommon that the more granular DSM-5 SUD diagnostic information is recorded by a clinician in an unstructured note. To date, many traditional natural language processing (NLP) efforts have been tested and implemented to identify additional patient risk factors in unstructured clinical notes, with varying degrees of success \cite{poulsen2022classifying, patra2021extracting}.
However, the high variability in language found within clinical notes imposes significant limits on the accuracy of traditional techniques that rely on parsing rules to detect text patterns. Clinician typos, neologisms, abbreviations, and other linguistic variations hinder the effectiveness of these methods.
Deep learning techniques, on the contrary, have demonstrated impressive efficacy in extracting such information from complex unstructured clinical notes \cite{romanowski2023extracting, patra2021extracting}.
Nevertheless, as information extraction is a supervised NLP task, the requirement for large-scale annotated datasets of high quality for training these models to reach their full potential poses an inevitable bottleneck in novel real-world applications.
Recently, Large Language Models (LLMs) have emerged as a promising solution to this challenge, particularly due to their ability to ``learn" and adapt to diverse language patterns without the need for additional model training \cite{zhao2023survey}.
LLMs are deep learning models in natural language processing equipped with millions or billions of parameters.
Although primarily trained on open-source, non-domain-specific texts, some studies have underscored their effectiveness when applied to clinical notes \cite{zhou2023exploring, guevara2024large, peng2023model, peng2023generative, alsentzer2023zero, agrawal2022large, choi2023developing}.

In this study, we investigate the utilization of LLMs to address the challenge of limited annotated datasets in a new clinical application---specifically, extracting information on the diagnosis severity specifiers of various SUD diagnoses from clinical notes.
Our proposed workflow involves performing zero-shot learning (ZSL) of LLMs by carefully crafting prompts and post-processing the generated text to extract SUD severity-related information.
In a zero-shot learning scenario, the LLM is prompted to execute a specific task via an input prompt, generating text in response, which serves as the output. For instance, when provided with the prompt: \textit{Extract the reference to alcohol use disorder diagnosis with surrounding information relevant to it from the diagnoses section in the following note. If you can't find the answer, please respond ``unanswerable". Note: $<$clinical note text$>$}. Then, the LLM produces the extracted text---\textit{severe etoh use d/o}---from the clinical note.
Leveraging Flan-T5 \cite{chung2022scaling}, an open-source LLM, we demonstrate through experimentation and comparative analysis that the LLM achieves better recall compared to rule-based regular expressions, highlighting its efficacy in this downstream task without requiring additional training.
For evaluation, we use a set of 577 notes annotated by a subject-matter expert (SME) who is a Licensed Clinical Psychologist. 
We focus on 11 categories of SUD diagnoses: alcohol; opioids; cannabis; sedatives, hypnotics/anxiolytics; cocaine; amphetamines; caffeine; hallucinogens; nicotine; inhalants; and other psychoactive substances.
To the best of our knowledge, no previous study has employed LLMs to extract information on severity specifiers of SUD diagnoses across 11 SUD categories.
This study is an initial step in our larger objective to devise an efficient approach to use LLMs to extract an array of nuanced information on SUD diagnoses, including severity specifiers, withdrawal symptoms, and social determinants of health, thereby contributing significantly to risk assessment, treatment planning, patient safety, recovery, and overall well-being.


\section*{Background}
\vspace{-.5em}
Large language models like GPT-4 \cite{achiam2023gpt} have revolutionized NLP by demonstrating a remarkable ability to interpret and generate human-like text across a wide range of domains \cite{zhao2023survey}. Unlike traditional deep learning models that require extensive domain-specific training data to achieve high performance, LLMs leverage their vast pre-training on diverse text to interpret context, make inferences, and generate relevant responses. Two techniques, namely zero-shot learning (ZSL) and few-shot learning (FSL), play pivotal roles in the context of LLMs. In both methods, the LLM performs a novel task on unseen data guided by a user-defined task description (prompt). The crucial distinction between ZSL and FSL is the extent of exposure the LLM receives during inference. ZSL entails no exposure, while FSL involves exposure to a limited number of examples. This is particularly valuable in the clinical domain, where acquiring expert-annotated data is both costly and time-consuming. 

Recent studies have explored the application of LLMs in the clinical domain\cite{zhou2023exploring, guevara2024large, peng2023model, peng2023generative}.
Alsentzer et al.\cite{alsentzer2023zero} used ZSL with the Flan-T5 model to extract postpartum hemorrhage (PPH) concepts from obstetric discharge summaries for interpretable phenotypes, achieving a high positive predictive value and identifying 45\% more patients with PPH than claims codes. 
Agrawal et al.\cite{agrawal2022large} used ZSL and FSL with GPT-3 based InstructGPT model\cite{ouyang2022training} to perform five tasks in the clinical domain: clinical sense disambiguation, biomedical evidence extraction, co-reference resolution, medication status extraction, and medication attribute extraction. Their findings demonstrated that GPT-3 achieves better performance than established baselines using guided handcrafted prompts.
\textit{HealthPrompt}\cite{sivarajkumar2022healthprompt} is a ZSL paradigm introduced to classify new classes without prior training data. The authors utilized ZSL to enable prompt-based learning by adjusting task definitions via 2 types of prompt templates: cloze prompts and prefix prompts. Using various pre-trained models and custom prompt templates, they demonstrated improved performance, particularly notable with ClinicalBERT \cite{alsentzer2019publicly}, which was trained on clinical texts.
In another study\cite{choi2023developing}, custom prompts were used to extract information from breast cancer pathology and ultrasound reports. They developed 12 prompts for extracting and standardizing data, finding that pre-trained language models, such as those from OpenAI, enabled efficient and cost-effective information extraction.

Numerous studies have focused on extracting targeted information related to substance use, employing rule-based techniques or a combination of rule-based and machine learning methods across various substances, including opioids \cite{poulsen2022classifying, poulsen2024developing, zhu2022automatically}, alcohol, tobacco, and drugs \cite{lybarger2021annotating, 10.1093/jamia/ocad012, 10.1093/jamia/ocad107, richie2023extracting, lybarger2018using}, and cannabis \cite{tavabi2023disparities, sajdeya2023developing}. These studies underscore the need for traditional manual annotation. Recent research, exemplified by \cite{guevara2024large, bhate2023zero, ramachandran2023prompt}, has employed ZSL and FSL prompting techniques within LLMs such as GPT-3.5, GPT-4, and Flan-T5 to extract SDoH information from clinical notes, marking a shift in approach towards data annotation challenges.

\section*{Methods}
\begin{figure}[H]
\vspace{-.5em}
  \centering
    \includegraphics[width=1.0\textwidth]{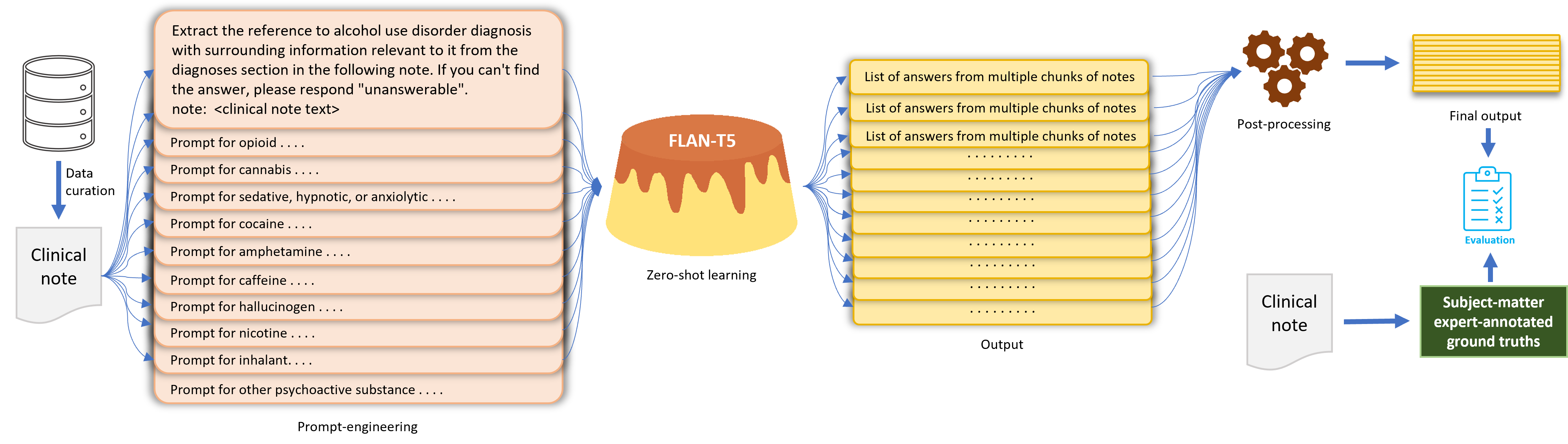}
\vspace{-1.5em}
 \caption{Overall workflow of the ZSL approach with a large language model FLan-T5. The steps in the workflow include data curation, data annotation, prompt engineering, zero-shot learning, post-processing of the model-generated output text, and performance evaluation.}
 \label{fig:workflow}
\end{figure}

In this section, we describe our dataset, the ZSL approach, and the experimental setup for using LLMs to extract information on SUD severity specifiers. We also briefly outline performance evaluation metrics. Figure \ref{fig:workflow} shows the overall workflow of this study, including data curation, annotation, information extraction using an LLM, post-processing, and evaluation.

\paragraph{Dataset}
For the study, we used a random set of 577 fully identified clinical notes belonging to 574 unique patients, curated from the United States Department of Veterans Affairs Corporate Data Warehouse (VA CDW).
While sampling, we ensured that the selected notes were associated with any of the following ICD-10 codes in the structured data---$F10,\ F11,\ F12,\ F13,\ F14,\ F15,\ F16,\ F17,\ F18,\ F19$---to increase their chance of containing information on SUD. Additionally, we augmented this selection with the following structured data criteria: admittance to a SUD clinic, presence of a CPT code of interest, but not any notes related to group therapy or drug test results. 
The notes were taken during patient visits occurring between October 2015 and November 2023.
The notes vary in length from 11 to 10719 tokens, with an average of 1043 tokens and a median of 423 tokens.
The types of notes comprise 79 varieties, including but not limited to mental health, suicide prevention, substance use disorder treatment program, education, psychiatry, administrative, psychology, and nursing.

An SME manually annotated each note to establish ground truths for diagnosis severity specifiers across 11 categories of substance use disorder diagnoses.
Observations reveal that the information about the severity level of diagnosed SUD is usually found in the section(s) of the note that mentions the current diagnoses during that patient visit.
We used this prior knowledge to build the prompts for the LLM later in the study.
During our post-annotation exploratory data analysis, we noted a highly imbalanced ratio between notes containing and those lacking information for each of the SUD categories, ranging from a minimum of 3\% to a maximum of 33\%, with an average of 10\%, as anticipated.

Ethics: This project was conducted as a national quality improvement effort to improve care for Veterans with substance use being treated in the Veterans Health Administration (VHA). Models were designed to be implemented into VHA decision support systems, and are not expected to be generalizable or valid for application outside of notes from the VHA Computerized Patient Record System (CPRS). As such, this work is considered non-research by VHA (as per ProgramGuide-1200-21-VHA-Operations-Activities.pdf (va.gov)). However, Oak Ridge National Laboratory (ORNL) required additional oversight of this VHA clinical quality improvement project as local standard practice for all uses of patient medical record data within their institution, with approval of the project by the ORNL IRB.



\paragraph{Zero-shot Learning Using a Large Language Model}
In light of the high performance in clinical concept extraction from unseen text passages without further training demonstrated in recent studies \cite{alsentzer2023zero, guevara2024large}, we opted for the Google Flan-T5 models \cite{chung2022scaling} for our experiments.
Flan-T5 is a family of large language models with encoder-decoder architecture, which was instruction-finetuned from Google T5 models \cite{raffel2020exploring} on 473 NLP datasets for 1836 NLP tasks utilizing a diverse range of instruction templates.
There are five models in the Flan-T5 family: Flan-T5-XXL, Flan-T5-XL, Flan-T5-Large, Flan-T5-Base, and Flan-T5-Small with 11B, 3B, 780M, 250M, and 80M parameters, respectively.
In the context of zero-shot learning, the Flan-T5 model takes a prompt outlining the task as input and generates text as output in response to the prompt (Figure \ref{fig:workflow}).
In this study, we leveraged Flan-T5 models to extract text indicating any substance use diagnosis severity specifiers from clinical notes using zero-shot learning.
To develop prompts for the task, we used a small development set containing 57 randomly selected clinical notes of 10\% of the patients within the dataset.
The remainder of the clinical notes were considered as the test set.
Note that there is no patient overlap between the development and test sets to avoid data leakage.

For prompt engineering, we drew inspiration from prior studies performing inference via zero-shot learning using Flan-T5 for clinical concept extraction \cite{alsentzer2023zero, choi2023developing}.
In our use case, we found that the model often fails when presented with straightforward prompts, such as, ``Extract the severity level of alcohol use disorder diagnosis in the following note. Note: $<$report text$>$".
To expedite progress, we employed a trial-and-error strategy based on prior knowledge acquired during data annotation. Specifically, our prompts are based on observing that the SUD severity specifier tends to be located adjacent to wherever the diagnosis is written in the note. Beginning with ``alcohol use disorder", we refined the prompt up to 10 iterations. Subsequently, we performed minimal prompt adjustments for the remaining SUD categories.
We provided the developed prompts for all SUD categories in our GitHub repository (available at: \url{https://github.com/mmahbub/SUDSeverity-Extract-LLM}).
Crucially, the location of a diagnosis in a clinical note is not based on any standardized note structure or format. Note structure widely varies based on providers, VA centers, locations, note types, and timeframes.
Hence, we deferred to the LLM to locate the diagnoses within the note and extract information from it.


Clinical notes frequently surpass the maximum allowable sequence length of 512 tokens in the input text for Flan-T5 models. This challenge is compounded when prompts are added to the beginning of the clinical notes.
To address this issue, we adopted a known technique in question-answering modeling: sliding window with a document stride \cite{devlin2018bert}.
In this technique, if the number of tokens in the input prompt is 50 and the number of tokens in the clinical notes exceeds 462, a sliding window with a document stride is applied to create chunks of the clinical note where each chunk consists of 462 tokens.
We used 128 as the document stride based on previous studies \cite{devlin2018bert}. Once the chunks are formed, the tokens from the prompt are prepended to each note chunk forming multiple input sequences of 512 tokens for the same note.
Note that, the length of the clinical notes and the maximum sequence length also restrict the few-shot inference capability of the Flan-T5 models. 

Considering that model-generated output text may contain irrelevant information as multiple chunks of the same note are presented to the model, we performed some post-processing steps for each SUD category from an operational point-of-view, as follows: 
(i) We removed answers that did not contain any of the phrases/substance names associated with the SUD category of interest.
(ii) Additionally, we filtered out answers lacking any of the specified phrases representing ``use disorder", such as ``dependence" or ``use d/o", following the substance names. The list of substance names is available in our GitHub repository along with the specified phrases denoting a use disorder.
(iii) Furthermore, to increase the reliability of the generated text and reduce hallucination in the final output, we filtered out the answers if any of the common substrings of length five or higher (measured in characters) between the note and the answer do not satisfy the conditions explained in (i) and (ii).
We chose a length of five for common substrings, determined by the shortest reference to substance use disorder in our dataset: ``mj ud" (where mj and ud represent marijuana and use disorder, respectively). 
The post-processing steps were only performed on the answers that did not contain the string ``unanswerable". Any filtered answers were replaced with ``unanswerable".

For our study, we conducted experiments with all variants of the Flan-T5 models and presented the evaluation scores for the best-performing one. We used the models from the HuggingFace \cite{wolf2019huggingface} model hub (available at: \url{https://huggingface.co/docs/transformers/main/en/model_doc/flan-t5}). For output text generation, we used the greedy decoding approach, setting the temperature to 1 and the maximum number of new tokens to 100.  
All our experiments were performed on a Linux virtual machine with two GRID V100-32C GPUs.

\paragraph{Information Extraction Using Regular Expressions}
We also performed information extraction using primitive rule-based regular expressions (RegEx). We created the rules following common patterns and using the phrases/substance names associated with the SUD categories, the terminology indicating a use disorder, and the phrases denoting the severity level of SUD diagnosis. The complete set of rules is available in our GitHub repository. 
We compared the performance of the Flan-T5 model to the RegEx approach for each SUD category.
Regular expressions serve as a fundamental benchmark since they, similar to the zero-shot learning of large language models, do not necessitate training.

\paragraph{Evaluation}
We measured the performance of the LLMs using two string matching criteria--strict and relaxed. As the names suggest, the strict match looks for character-by-character matches between the ground truth and the generated text while the relaxed match just looks for overlap. Based on these criteria, we used the F1, precision, and recall scores to assess the performance of the LLMs.  
In this scenario, True Positive denotes the number of tokens that are common between the ground truth and the generated answer, False Negative indicates the number of tokens present solely in the ground truth but absent in the generated answer, and False Positive represents the count of tokens exclusively found in the generated answer.
The relaxed F1, precision, or recall scores per sample have a range from 0 to 1. As per a prior study \cite{rajpurkar-etal-2016-squad}, we present the macro-averaged F1, precision, and recall scores, on the test sets.


\section*{Results}
\vspace{-.5em}
We report the performance of the best-performing Flan-T5 model, Flan-T5-XXL, in the zero-shot learning paradigm, specifically tailored to our task: extracting information for SUD diagnosis severity specifiers.
The Flan-T5-XXL model outperforms other models in the Flan-T5 family in every evaluation metric across all SUD categories. This finding aligns with existing literature, indicating that larger models excel in zero-shot learning. Furthermore, it underscores the generalization capability of these models when presented with previously unseen data without requiring additional training.
The inference is performed on a random selection of 520 annotated notes.
In Table \ref{tab:llm-result}, for each SUD category, we report the average F1 score for the strict match and F1, precision, and recall scores for the relaxed match achieved by the Flan-T5-XXL model. 

\begin{table}[!htbp]
\centering
\vspace{.2em}
\caption{Performance scores of Flan-T5-XXL in extracting information on severity across 11 SUD categories}
\vspace{-.8em}
\resizebox{\columnwidth}{!}{%
\begin{tabular}{|l|cccc|c|c|}
\hline
\multirow{2}{*}{Substance Use Disorder Category} &
  \multicolumn{4}{c|}{With any SUD information} &
  Without any SUD information &
  Combined \\ \cline{2-7} 
 &
  \multicolumn{1}{c|}{F1 (Strict)} &
  \multicolumn{1}{c|}{F1 (Relaxed)} &
  \multicolumn{1}{c|}{Recall (Relaxed)} &
  Precision (Relaxed) &
  F1 (Strict) &
  F1 (Strict) \\ \hline
Alcohol &
  \multicolumn{1}{c|}{61.24} &
  \multicolumn{1}{c|}{83.98} &
  \multicolumn{1}{c|}{88.52} &
  84.36 &
  93.86 &
  82.69 \\ \hline
Opioid &
  \multicolumn{1}{c|}{56.16} &
  \multicolumn{1}{c|}{77.55} &
  \multicolumn{1}{c|}{82.91} &
  77.36 &
  98.21 &
  92.31 \\ \hline
Cannabis &
  \multicolumn{1}{c|}{51.28} &
  \multicolumn{1}{c|}{71.23} &
  \multicolumn{1}{c|}{76.41} &
  72.37 &
  97.29 &
  90.38 \\ \hline
Sedative, hypnotic, or anxiolytic &
  \multicolumn{1}{c|}{51.61} &
  \multicolumn{1}{c|}{69.11} &
  \multicolumn{1}{c|}{70.10} &
  70.76 &
  99.59 &
  96.73 \\ \hline
Cocaine &
  \multicolumn{1}{c|}{57.89} &
  \multicolumn{1}{c|}{80.88} &
  \multicolumn{1}{c|}{85.72} &
  82.29 &
  92.44 &
  88.65 \\ \hline
Amphetamine &
  \multicolumn{1}{c|}{53.49} &
  \multicolumn{1}{c|}{72.81} &
  \multicolumn{1}{c|}{77.21} &
  73.66 &
  95.81 &
  92.31 \\ \hline
Caffeine &
  \multicolumn{1}{c|}{53.33} &
  \multicolumn{1}{c|}{59.05} &
  \multicolumn{1}{c|}{58.33} &
  60.00 &
  98.81 &
  97.50 \\ \hline
Hallucinogen &
  \multicolumn{1}{c|}{52.38} &
  \multicolumn{1}{c|}{63.57} &
  \multicolumn{1}{c|}{71.43} &
  61.70 &
  100.00 &
  98.08 \\ \hline
Nicotine &
  \multicolumn{1}{c|}{68.42} &
  \multicolumn{1}{c|}{80.26} &
  \multicolumn{1}{c|}{83.60} &
  80.27 &
  97.41 &
  94.23 \\ \hline
Inhalant &
  \multicolumn{1}{c|}{66.67} &
  \multicolumn{1}{c|}{82.98} &
  \multicolumn{1}{c|}{93.06} &
  80.85 &
  100.00 &
  98.85 \\ \hline
Other psychoactive substance &
  \multicolumn{1}{c|}{45.00} &
  \multicolumn{1}{c|}{66.98} &
  \multicolumn{1}{c|}{68.75} &
  68.67 &
  100.00 &
  97.88 \\ \hline
\end{tabular}%
}
\label{tab:llm-result}
\vspace{-.5em}

\end{table}

By adopting the sliding window approach to handle the length of clinical notes, it is possible to obtain multiple candidate answers from more than one chunk of one clinical note.
To refine these answers, our post-processing steps filter out the irrelevant answers.
For those remaining notes with multiple candidate answers, we select the most suitable one based on the F1 score for the relaxed match.
However, when ground truth data is absent in an operational setting, determining the correct answer from a set of candidates may impose an additional time overhead.
In contrast to a previous study \cite{alsentzer2023zero} that used a union of candidate answer strings to derive the final answer, this method seemed non-viable for our specific application. Consequently, the development of a more intricate strategy for handling such scenarios is imperative and warrants thorough exploration in future research endeavors.



\begin{wrapfigure}{p}{0.35\textwidth}
\centering
\includegraphics[scale=0.43]{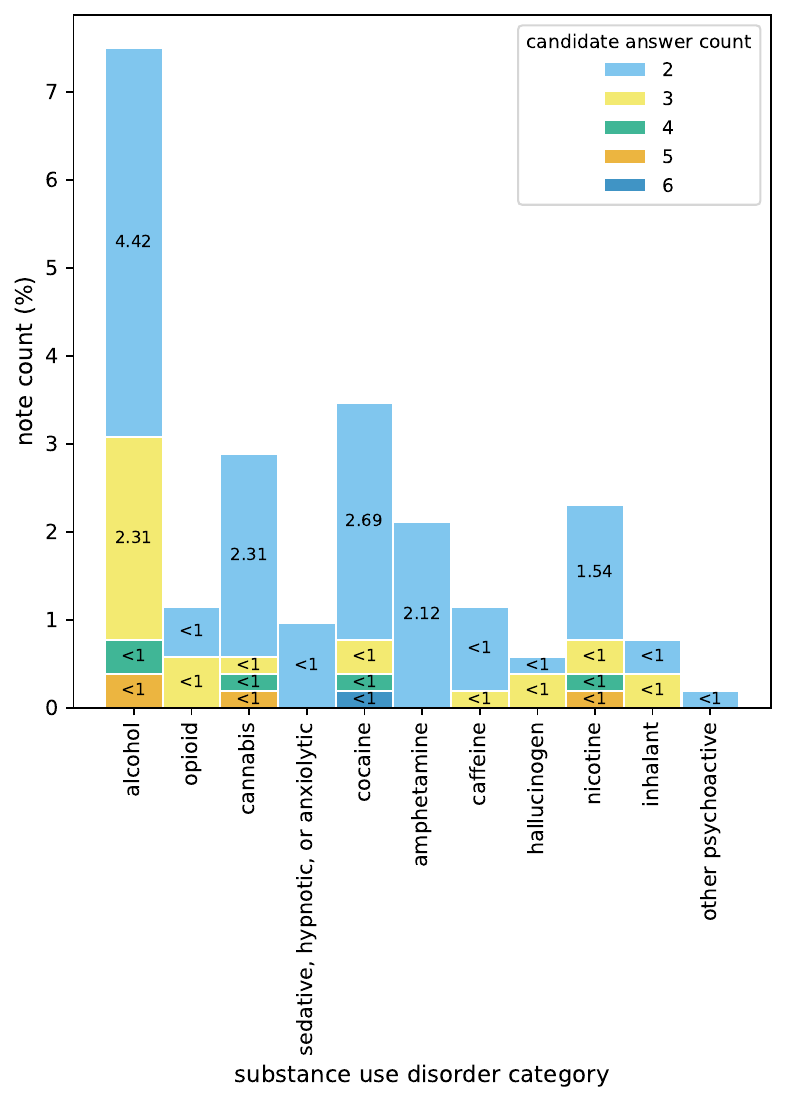}
\caption{\label{fig:error-candans}Distribution of candidate answer count (higher than 1)}
\end{wrapfigure}

Figure \ref{fig:error-candans} shows the distribution of the number of candidate answers for notes with more than one candidate.
Interestingly, very few of the notes have more than one candidate answer. We also find that for all SUD categories, there is either a significantly low or insignificant correlation between the length of the notes (measured by counting tokens) and the number of candidate answers---evidenced by Pearson correlation coefficients at the significance levels of 0.05, 0.01, and 0.001, as detailed in Table \ref{tab:corr}.

\vspace{.4em}
\begin{table}[!htbp]
\centering
\caption{Pearson correlation coefficient to measure the correlation between the length of the notes (in the number of tokens) and the number of candidate answers. The significance of the correlation is measured at significance levels of 0.05*, 0.01**, and 0.001***}
\vspace{-.8em}
\resizebox{\columnwidth}{!}{%
\begin{tabular}{|m{2.5cm}|l|l|l|m{2cm}|l|l|l|l|l|l|m{1.4cm}|}
\hline
substance use disorder category &
  alcohol &
  opioid &
  cannabis &
  sedative, hypnotic, or anxiolytic &
  cocaine &
  amphetamine &
  caffeine &
  hallucinogen &
  nicotine &
  inhalant &
  other psychoactive \\ \hline
Pearson correlation coefficient &
  0.44*** &
  0.11** &
  0.26*** &
  0.12** &
  0.31*** &
  0.19*** &
  0.22*** &
  0.16*** &
  0.29*** &
  0.17*** &
  0.06 \\ \hline
\end{tabular}%
}
\vspace{-.8em}
\label{tab:corr}
\end{table}

\begin{figure}[!htbp]
    \centering
    \vspace{-.8em}
    \includegraphics[width=.7\textwidth]{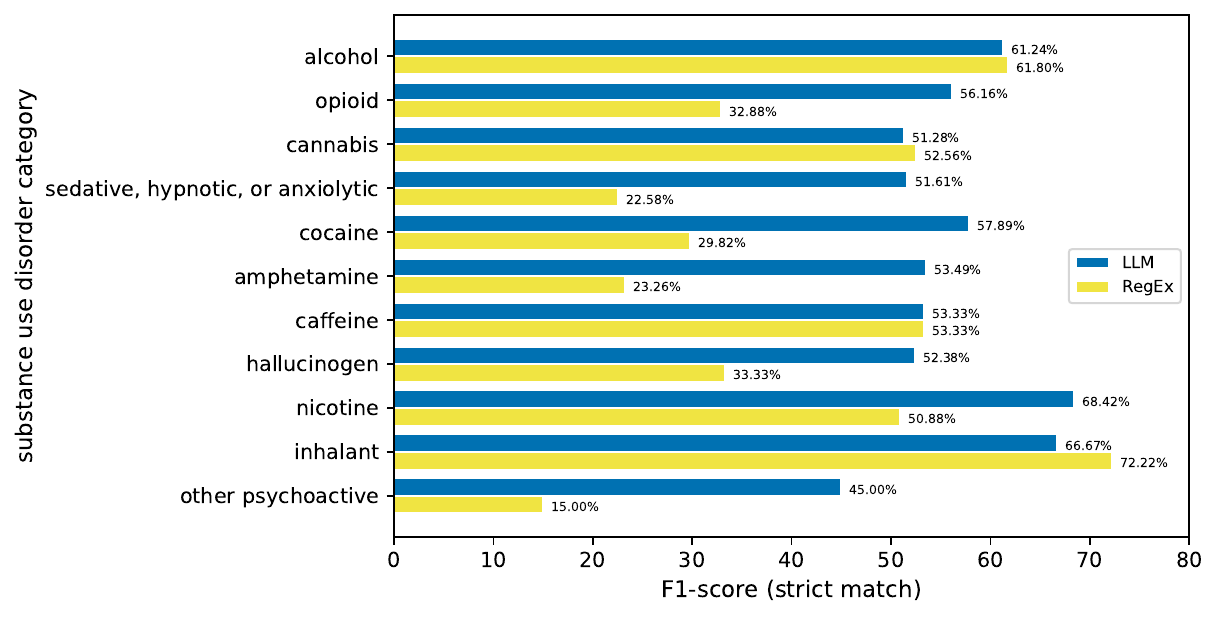}
    \vspace{-.8em}
    \caption{Performance comparison (in strict match F1-score) between LLM and RegEx across all SUD categories}
    \label{fig:rule-llm-compare}
\end{figure}

In Figure \ref{fig:rule-llm-compare}, we compare the performance of the Flan-T5 model and RegEx across all SUD categories, evaluating F1 scores for strict matches.
The findings indicate that the LLM outperforms the RegEx approach in seven out of 11 SUD categories.
This difference in performance can be attributed to the nuanced and diverse expressions of SUD diagnoses in clinical notes, which do not align well with the rigid nature of RegEx. 
Furthermore, the better performance of RegEx over LLM in the four remaining categories (demonstrated in Figure \ref{fig:rule-llm-compare}) sheds light on how some SUD diagnoses in clinical notes may be comparatively more structured than others. For instance, SUDs can be recorded as causing a secondary disorder (e.g., stimulant-induced anxiety disorder) and increase variability in diagnosis presentations. Conversely, certain substances such as caffeine, cannabis, and hallucinogens are indicated as not being able to induce certain types of secondary disorders\cite{DSM}, thus reducing the variability in diagnosis presentations. Additionally, these substance categories, as well as alcohol, present less variation in the name of the substance in the diagnosis. It may be that RegEx performs better on average when there is less variation in the substance's name, e.g., alcohol diagnoses are likely to have either ``alcohol" or ``ETOH", whereas a diagnosis for opioids may have a range of parentheticals, like ``opioid SUD (kratom)", due to the variety of opioids a person may be using.
Additionally, when creating the RegEx rules, we could not omit ``rule out" diagnoses, i.e. diagnoses to be further assessed but not assigned to the patient at the visit. The possible presence of such notes may also affect the apparent performance of RegEx.

\begin{figure}[!htbp]
\vspace{-.8em}
    \centering
    \includegraphics[width=.8\textwidth]{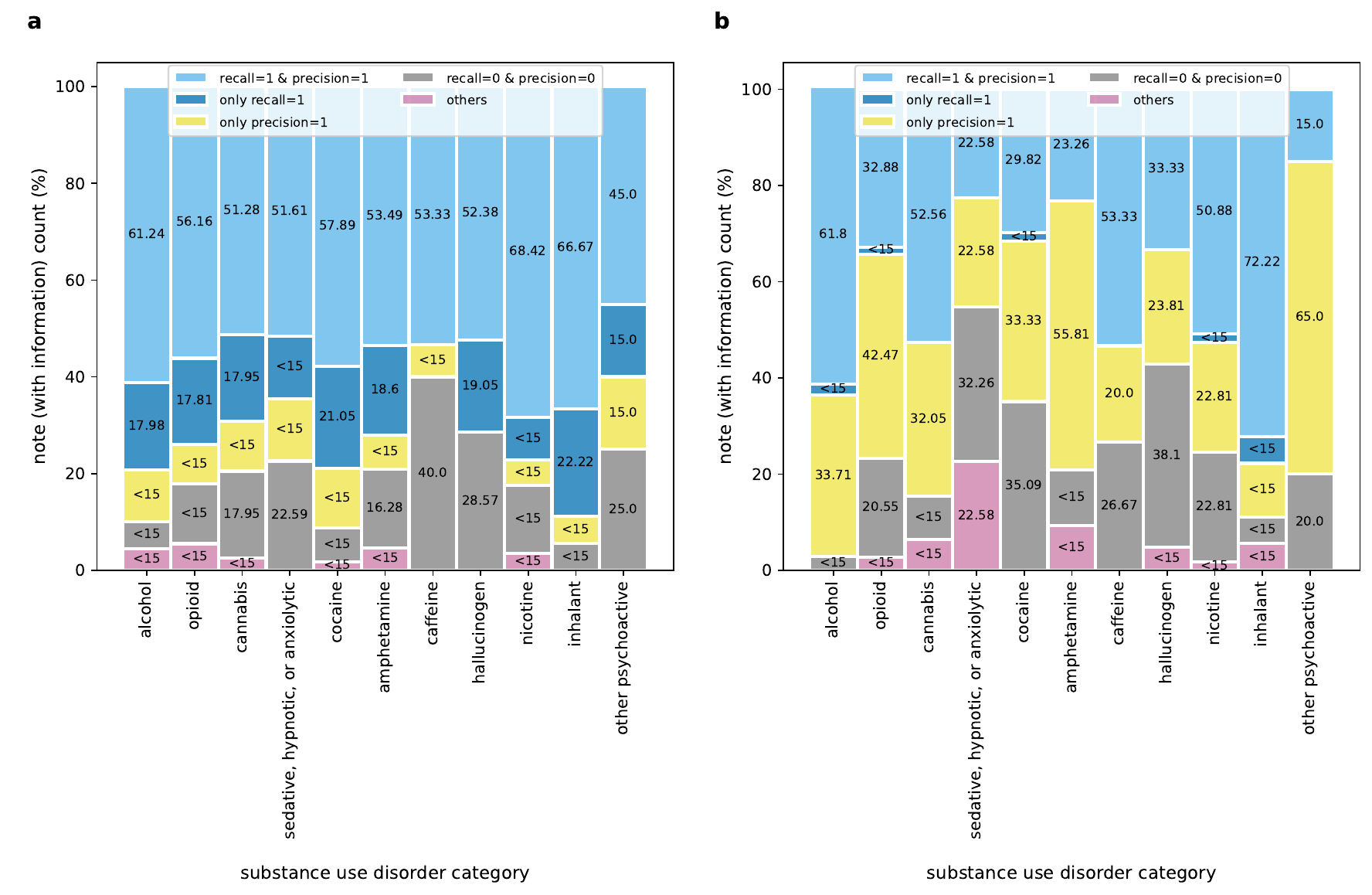}
    \vspace{-.8em}
    \caption{Comprehensive evaluation of the performance of LLM (\textbf{a}) and RegEx (\textbf{b}) for notes containing any SUD information}
    \label{fig:error-perf}
\end{figure}

In Figure \ref{fig:error-perf}, we provide a detailed analysis of performance metrics of notes containing any mention of substance use disorders.
This analysis offers a more comprehensive overview of the LLM's efficacy in extracting information on SUD diagnosis severity.
To pinpoint the nature of errors, we categorize instances based on whether the answers precisely match the ground truth or not.
Specifically, we scrutinize instances where either the recall score equals 1 or the precision score equals 1, but the strict match F1-score equals 0.
Additionally, we examine cases where the LLM completely fails to extract any SUD information, characterized by both recall and precision scores of 0.
A notable observation from Figure \ref{fig:error-perf}a is that in a significant portion of notes where the LLM fails to strictly extract ground truth answers, it successfully extracts strings fully containing the ground truth (recall=1). On the contrary, fewer instances with precision scores of 1 indicate that the LLM tends to generate extra words during task execution, which is an anticipated behavior.
Figure \ref{fig:error-perf}b highlights the opposite characteristics of the rule-based RegEx approach. As a more rigid string search technique, RegEx excels at capturing some parts of the ground truths---indicated by the precision scores, as opposed to finding strings containing the ground truth---depicted by the recall scores.
Despite the inferior performance of RegEx compared to the LLM, further exploratory analysis reveals instances where RegEx outperforms the LLM in capturing information in clinical notes, particularly for SUD categories such as alcohol, cannabis, caffeine, and inhalants. This phenomenon could be attributed to specific note structures and should be investigated further in future studies.


\section*{Discussion and Conclusion}
\vspace{-.5em}
LLMs offer significant promise in enhancing clinical practice by facilitating the extraction of crucial information from lengthy and intricate clinical documents.
Unlike conventional deep learning models, LLMs often do not necessitate supplementary training to adjust to novel tasks, thus alleviating the bottleneck of annotating large quantities of high-quality data for new downstream tasks.
In this study, we address the challenge of determining SUD severity specifiers from clinical notes---a daunting task owing to the variations in how they are written in clinical notes. Examples of this variation that indicate the difficulty in using strictly rule-based methods (including variations in spelling, abbreviations, and syntax) are: 
\vspace{-1em}
\begin{multicols}{2}
\begin{itemize}
\setlength\itemsep{-6pt}
    \item ``cannabis/alcohol/opioid use disorder: mild"
    \item ``alcohol use disorder mod/severe"
    \item ``moderate caffeine use do"
    \item ``opioid (heroin/ vicodin) use disorder - severe (on agonist therapy)"
    \item ``marijuana user (in remission)"
    \item ``cannabis (thc vape) use disorder, mild"
    \item ``(ephedrine) sedative use disorder, in sustained remission"
    \item ``sedative hypnotic use disorder, severe (xanax)"
    \item ``meets criteria for substance use disorder: cocaine [] mild (2-3); [] moderate (4-5); [x] severe (6 or more)"
    \item ``amphetamine (methamphetamine) or other stimulant, without perceptual disturbances disorder, sever, in remission"
    \item ``other hallucinogen use disorder (mdma/ecstasy), moderate, in remission"
    \item ``other/unknown substance disorder: severe (coricidin)"
    \item ``moderate inhalant (nitrous oxide) use d/o"
\end{itemize}
\end{multicols}
\vspace{-1.4em}


This study is the first foundational effort in a broader goal to extract additional granular risk factors associated with SUD patients.
These include but are not limited to the SUD severity specifiers, chronology, co-determinants (e.g., withdrawal symptoms), and social determinants of health---often excluded from the structured data in the electronic health records.
This detailed information plays a vital role in risk assessment, guiding effective treatment plans, ensuring patient safety, and fostering recovery and overall well-being.
We began this effort with SUD severity specifiers in the clinical notes as a proof of concept in part because such specifiers are clearly indicated and described in the DSM-5 (e.g., mild, moderate, and severe)\cite{DSM}. Yet, even with such clear documentation for these specifiers, our work has shown that additional research is necessary with both the LLM and rule-based approaches to achieve high levels of accuracy. In this vein, other risk factors that are less clearly documented (social determinants, for instance) are likely to pose additional challenges.
To the best of our knowledge, no prior studies have specifically targeted the extraction of severity specifiers across 11 distinct categories of SUD diagnoses from clinical notes.

Our findings indicate that, when provided with appropriate prompts, the Flan-T5 model excels in extracting information with high recall and precision across most SUD categories.
Particularly, it outperforms RegEx in extracting nuanced information that cannot be easily captured by rigid rules.
Nonetheless, the LLM occasionally struggles to extract information accurately and succinctly, suggesting that an ensemble approach combining both RegEx and LLM may be beneficial in such instances.
Moreover, the errors observed in the study indicate that finetuning the LLM based on specific instructions could enhance overall performance.

There are several key limitations in this study.
Firstly, the evaluation was conducted on a relatively small test dataset comprising 520 annotated notes, with even fewer containing the target information.
Expanding this annotated dataset necessitates significant manual effort from SMEs.
As a future direction, we aim to explore more efficient annotation methods, potentially using a hybrid approach that combines LLMs and RegEx to assist SMEs.
Secondly, our study exclusively utilized Flan-T5 models.
Future research could benefit from experimenting with LLMs of diverse architectures, pre-training paradigms, pre-training datasets, and higher model parameters, such as LLAMA2-13B, LLAMA2-70B, or Mistral-7B.
Also, we could not use closed models such as GPT-4 due to privacy concerns and data-use restrictions.
Thirdly, exploring instruction-finetuning to enhance the LLM's performance for this specific task is subjected to further efforts.
Additionally, it is important to acknowledge that Flan-T5 models, similar to their counterparts, are susceptible to hallucinations and bias.
While our post-processing methods aim to mitigate hallucinations to some extent, careful consideration and attention are essential when considering the utilization of such models in critical and sensitive clinical applications.


\section*{Acknowledgments}
This work was supported by Department of Veterans Affairs, Office of Mental Health and Suicide Prevention. This research used resources of the Knowledge Discovery Infrastructure at the Oak Ridge National Laboratory, which is supported by the Office of Science of the U.S. Department of Energy under Contract No. DE-AC05-00OR22725 and the Department of Veterans Affairs Office of Health Informatics and by VA-DoD Joint Incentive fund under IAA No. 36C10B21M0005.
The authors wish to acknowledge the support of the larger partnership. Most importantly, the authors would like to thank and acknowledge the veterans who chose to get their care at the VA.
\textit{Disclaimer:} The views and opinions expressed in this manuscript are those of the authors and do not represent those of the Department of Veterans Affairs, the Department of Energy, or the United States Government.

$^*$These authors contributed equally to the development of the paper.

\makeatletter
\renewcommand{\@biblabel}[1]{\hfill #1. }
\makeatother

\bibliographystyle{vancouver}
\bibliography{amia}  

\begin{thebibliography}{10}

\bibitem{ICD}
World~Health Organization.
\newblock The ICD-10 classification of mental and behavioural disorders:
  clinical descriptions and diagnostic guidelines. vol.~1.
\newblock World Health Organization; 1992.

\bibitem{DSM}
American~Psychiatric Association.
\newblock Diagnostic and statistical manual of mental disorders: DSM-5.
\newblock 5th ed. American Psychiatric Association; 2013.

\bibitem{poulsen2022classifying}
Poulsen MN, Freda PJ, Troiani V, Davoudi A, Mowery DL.
\newblock Classifying characteristics of opioid use disorder from hospital
  discharge summaries using natural language processing.
\newblock Frontiers in Public Health. 2022;10:850619.

\bibitem{patra2021extracting}
Patra BG, Sharma MM, Vekaria V, Adekkanattu P, Patterson OV, Glicksberg B,
  et~al.
\newblock Extracting social determinants of health from electronic health
  records using natural language processing: a systematic review.
\newblock Journal of the American Medical Informatics Association.
  2021;28(12):2716-27.

\bibitem{romanowski2023extracting}
Romanowski B, Ben~Abacha A, Fan Y.
\newblock Extracting social determinants of health from clinical note text with
  classification and sequence-to-sequence approaches.
\newblock Journal of the American Medical Informatics Association.
  2023:ocad071.

\bibitem{zhao2023survey}
Zhao WX, Zhou K, Li J, Tang T, Wang X, Hou Y, et~al.
\newblock A survey of large language models.
\newblock arXiv preprint arXiv:230318223. 2023.

\bibitem{zhou2023exploring}
Zhou J, Li T, Fong SJ, Dey N, Crespo RG.
\newblock Exploring chatGPT'S potential for consultation, recommendations and
  report diagnosis: Gastric cancer and gastroscopy reports’ case.
\newblock IJIMAI. 2023;8(2):7-13.

\bibitem{guevara2024large}
Guevara M, Chen S, Thomas S, Chaunzwa TL, Franco I, Kann BH, et~al.
\newblock Large language models to identify social determinants of health in
  electronic health records.
\newblock NPJ digital medicine. 2024;7(1):6.

\bibitem{peng2023model}
Peng C, Yang X, Smith KE, Yu Z, Chen A, Bian J, et~al.
\newblock Model Tuning or Prompt Tuning? A Study of Large Language Models for
  Clinical Concept and Relation Extraction.
\newblock arXiv preprint arXiv:231006239. 2023.

\bibitem{peng2023generative}
Peng C, Yang X, Chen A, Yu Z, Smith KE, Costa AB, et~al.
\newblock Generative Large Language Models Are All-purpose Text Analytics
  Engines: Text-to-text Learning Is All Your Need.
\newblock arXiv preprint arXiv:231206099. 2023.

\bibitem{alsentzer2023zero}
Alsentzer E, Rasmussen MJ, Fontoura R, Cull AL, Beaulieu-Jones B, Gray KJ,
  et~al.
\newblock Zero-shot Interpretable Phenotyping of Postpartum Hemorrhage Using
  Large Language Models.
\newblock medRxiv. 2023:2023-05.

\bibitem{agrawal2022large}
Agrawal M, Hegselmann S, Lang H, Kim Y, Sontag D.
\newblock Large language models are few-shot clinical information extractors.
\newblock In: Proceedings of the 2022 Conference on Empirical Methods in
  Natural Language Processing; 2022. p. 1998-2022.

\bibitem{choi2023developing}
Choi HS, Song JY, Shin KH, Chang JH, Jang BS.
\newblock Developing prompts from large language model for extracting clinical
  information from pathology and ultrasound reports in breast cancer.
\newblock Radiation Oncology Journal. 2023;41(3):209.

\bibitem{chung2022scaling}
Chung HW, Hou L, Longpre S, Zoph B, Tay Y, Fedus W, et~al.
\newblock Scaling instruction-finetuned language models.
\newblock arXiv preprint arXiv:221011416. 2022.

\bibitem{achiam2023gpt}
Achiam J, Adler S, Agarwal S, Ahmad L, Akkaya I, Aleman FL, et~al.
\newblock Gpt-4 technical report.
\newblock arXiv preprint arXiv:230308774. 2023.

\bibitem{ouyang2022training}
Ouyang L, Wu J, Jiang X, Almeida D, Wainwright C, Mishkin P, et~al.
\newblock Training language models to follow instructions with human feedback.
\newblock Advances in Neural Information Processing Systems. 2022;35:27730-44.

\bibitem{sivarajkumar2022healthprompt}
Sivarajkumar S, Wang Y.
\newblock Healthprompt: A zero-shot learning paradigm for clinical natural
  language processing.
\newblock In: AMIA Annual Symposium Proceedings. vol. 2022. American Medical
  Informatics Association; 2022. p. 972.

\bibitem{alsentzer2019publicly}
Alsentzer E, Murphy JR, Boag W, Weng WH, Jin D, Naumann T, et~al.
\newblock Publicly available clinical BERT embeddings.
\newblock arXiv preprint arXiv:190403323. 2019.

\bibitem{poulsen2024developing}
Poulsen MN, Freda PJ, Troiani V, Mowery DL.
\newblock Developing a Framework to Infer Opioid Use Disorder Severity From
  Clinical Notes to Inform Natural Language Processing Methods:
  Characterization Study.
\newblock JMIR Mental Health. 2024;11(1):e53366.

\bibitem{zhu2022automatically}
Zhu VJ, Lenert LA, Barth KS, Simpson KN, Li H, Kopscik M, et~al.
\newblock Automatically identifying opioid use disorder in non-cancer patients
  on chronic opioid therapy.
\newblock Health Informatics Journal. 2022;28(2):14604582221107808.

\bibitem{lybarger2021annotating}
Lybarger K, Ostendorf M, Yetisgen M.
\newblock Annotating social determinants of health using active learning, and
  characterizing determinants using neural event extraction.
\newblock Journal of Biomedical Informatics. 2021;113:103631.

\bibitem{10.1093/jamia/ocad012}
Lybarger K, Yetisgen M, Uzuner {\"O}.
\newblock {The 2022 n2c2/UW shared task on extracting social determinants of
  health}.
\newblock Journal of the American Medical Informatics Association. 2023
  04;30(8):1367-78.
\newblock Available from: \url{https://doi.org/10.1093/jamia/ocad012}.

\bibitem{10.1093/jamia/ocad107}
Peng C, Yang X, Yu Z, Bian J, Hogan WR, Wu Y.
\newblock {Clinical concept and relation extraction using prompt-based machine
  reading comprehension}.
\newblock Journal of the American Medical Informatics Association. 2023
  06;30(9):1486-93.
\newblock Available from: \url{https://doi.org/10.1093/jamia/ocad107}.

\bibitem{richie2023extracting}
Richie R, Ruiz VM, Han S, Shi L, Tsui F.
\newblock Extracting social determinants of health events with
  transformer-based multitask, multilabel named entity recognition.
\newblock Journal of the American Medical Informatics Association.
  2023:ocad046.

\bibitem{lybarger2018using}
Lybarger K, Yetisgen M, Ostendorf M.
\newblock Using neural multi-task learning to extract substance abuse
  information from clinical notes.
\newblock In: AMIA Annual Symposium Proceedings. vol. 2018. American Medical
  Informatics Association; 2018. p. 1395.

\bibitem{tavabi2023disparities}
Tavabi N, Raza M, Singh M, Golchin S, Singh H, Hogue GD, et~al.
\newblock Disparities in cannabis use and documentation in electronic health
  records among children and young adults.
\newblock NPJ digital medicine. 2023;6(1):138.

\bibitem{sajdeya2023developing}
Sajdeya R, Mardini MT, Tighe PJ, Ison RL, Bai C, Jugl S, et~al.
\newblock Developing and validating a natural language processing algorithm to
  extract preoperative cannabis use status documentation from unstructured
  narrative clinical notes.
\newblock Journal of the American Medical Informatics Association.
  2023:ocad080.

\bibitem{bhate2023zero}
Bhate NJ, Mittal A, He Z, Luo X.
\newblock Zero-shot Learning with Minimum Instruction to Extract Social
  Determinants and Family History from Clinical Notes using GPT Model.
\newblock In: 2023 IEEE International Conference on Big Data (BigData). IEEE;
  2023. p. 1476-80.

\bibitem{ramachandran2023prompt}
Ramachandran GK, Fu Y, Han B, Lybarger K, Dobbins NJ, Uzuner {\"O}, et~al.
\newblock Prompt-based Extraction of Social Determinants of Health Using
  Few-shot Learning.
\newblock arXiv preprint arXiv:230607170. 2023.

\bibitem{raffel2020exploring}
Raffel C, Shazeer N, Roberts A, Lee K, Narang S, Matena M, et~al.
\newblock Exploring the limits of transfer learning with a unified text-to-text
  transformer.
\newblock The Journal of Machine Learning Research. 2020;21(1):5485-551.

\bibitem{devlin2018bert}
Devlin J, Chang MW, Lee K, Toutanova K.
\newblock {BERT}: Pre-training of Deep Bidirectional Transformers for Language
  Understanding.
\newblock In: Proceedings of the 2019 Conference of the North {A}merican
  Chapter of the Association for Computational Linguistics: Human Language
  Technologies, Volume 1 (Long and Short Papers). Minneapolis, Minnesota:
  Association for Computational Linguistics; 2019. p. 4171-86.
\newblock Available from: \url{https://aclanthology.org/N19-1423}.

\bibitem{wolf2019huggingface}
Wolf T, Debut L, Sanh V, Chaumond J, Delangue C, Moi A, et~al.
\newblock Transformers: State-of-the-Art Natural Language Processing.
\newblock In: Proceedings of the 2020 Conference on Empirical Methods in
  Natural Language Processing: System Demonstrations. Online: Association for
  Computational Linguistics; 2020. p. 38-45.
\newblock Available from: \url{https://aclanthology.org/2020.emnlp-demos.6}.

\bibitem{rajpurkar-etal-2016-squad}
Rajpurkar P, Zhang J, Lopyrev K, Liang P.
\newblock {SQ}u{AD}: 100,000+ Questions for Machine Comprehension of Text.
\newblock In: Proceedings of the 2016 Conference on Empirical Methods in
  Natural Language Processing. Austin, Texas: Association for Computational
  Linguistics; 2016. p. 2383-92.
\newblock Available from: \url{https://aclanthology.org/D16-1264}.

\end{thebibliography}

\end{document}